# Stain-invariant representation for tissue classification in histology images

**Author**

Manahil Raza, Saad Bashir, Talha Qaiser, Nasir Rajpoot – Tissue Image Analytics Centre, Department of Computer Science, University of Warwick, Coventry CV4 7AL, U.K.

**Abstract**

The process of digitizing histology slides involves multiple factors that can affect a whole slide image's (WSI) final appearance, including the staining protocol, scanner, and tissue type. This variability constitutes a domain shift and results in significant problems when training and testing deep learning (DL) algorithms in multi-cohort settings. As such, developing robust and generalizable DL models in computational pathology (CPath) remains an open challenge. In this regard, we propose a framework that generates stain-augmented versions of the training images using stain matrix perturbation. Thereafter, we employed a stain regularization loss to enforce consistency between the feature representations of the source and augmented images. Doing so encourages the model to learn stain-invariant and consequently, domain-invariant feature representations. We evaluated the performance of the proposed model on cross-domain multi-class tissue type classification of colorectal cancer images and have achieved improved performance compared to other state-of-the-art methods.

**Introduction**

The advent of Deep Learning (DL) has revolutionised the field of Computational Pathology (CPath) and has enabled the automated and quantitative analysis of histology images [1,2]. Despite the success of DL methods, they are vulnerable to domain-specific variations [3]. Some major sources of variations include staining and scanning processes, where distinct institutions may employ different staining protocols and use various scanners, resulting in differences in the visual appearance of whole slide images (WSIs). This variability poses a significant challenge known as domain shift for training robust DL models and encumbers their ability to generalise well across diverse histology datasets. Addressing the domain shift problem has been a focal point for CPath researchers, leading to several efforts in stain normalisation [4-6], augmentation and adaptation. Stain augmentation (SA) aims to mitigate the effects of domain shift by generating augmented variations of the source images to mimic the stain variations present in the target domain for improving the model's generalisability on unseen data [7-9]. Tellez *et al.* [24] has stressed upon the importance of using stain augmentations for histopathology images for a more robust classification performance. Abbet *et al.* [11] proposed a novel domain adaptation (DA) method, Self-Rule to Multi-Adapt (SRMA), for single-source and muti-source tissue classification with multiple datasets by using in-domain and cross-domain losses. Unlike in DA, domain generalisation (DG) methods cannot leverage unlabelled data from the target domain [10]. To this effect, Vuong *et al.* [12] adopted a self-supervised contrastive learning approach using a combination of encoders and momentum encoders for colorectal cancer classification using patch shuffling augmentations. Our proposed method, inspired by [14] uses stain augmentations to help extract domain-invariant feature representations for colorectal cancer tissue images, thus ensuring that the class labels assigned to an image remain consistent in the face of staining variability.

**Methodology**

The proposed framework comprises of two modules, one for classification and the other for stain augmentation, as shown in Fig. 1. Each image *x* with label *y* is passed through ResNet-18 based feature extractor $f_e$ for extracting the feature representation as $f_e(x) = r$. This feature representation is then passed onto an MLP-based classifier $f_c$ for a classification decision, $\hat{y}$ as $f_c(f_e(x)) = \hat{y}$ During the training process, we also employ the stain augmentation network, which generates stain-altered version(s) of the source image as $x' = \{x'_1..x'_N\}$. For this purpose, we use the Vahadane [18] method for extracting the stain matrix. The stain concentrations are perturbed to create the stain-altered images using [15]. These images are then passed onto the same feature extractor $f_e(x') = r' = \{r'_1..r'_N\}$ to extract feature representations of the stain-altered images. Two loss functions are employed in the proposed workflow. We use the cross-entropy loss as the primary classification loss $L_c$ between the predicted label $\hat{y}$ and the ground-truth label *y*. Additionally, we employed a mean squared error (MSE) loss as a stain regularisation loss, $L_s = ||r - r'||_2^2$, which measures the distance between the extracted feature representations of the source and augmented images. The MSE loss acts as a strong penalisation factor, enforcing consistency in the face of stain augmentations. The overall loss function combines the two loss functions, $L = L_c + L_s$.

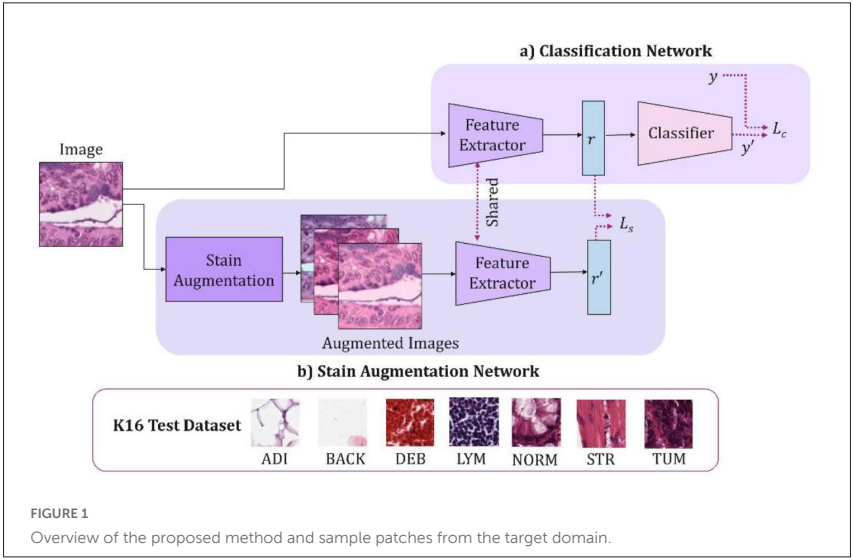

**FIGURE 1**
Overview of the proposed method and sample patches from the target domain.

**Results and Discussions**

We have employed two datasets to validate the proposed framework 1). Kather-19 (K19) [16], which contains 100,000 images (224×224 pixels) from 9 tissue classes and 2). Kather-16 (K16) [17], which consists of 5,000 images (150×150 pixels) from 8 classes. Since there are discrepancies between the class labels of the two datasets, we followed the strategy for relabelling [11] and grouped the data into seven classes, namely adipose, background, debris, lymphocytes, normal colon mucosa, stroma and colorectal adenocarcinoma epithelium. Sample images from K16 are shown in Fig.1. The results of the experiments are reported in Table 1, where ImageNet Upper Bound denotes an experiment where both the training and testing are performed with the same dataset (K16). The degradation in performance in ImageNet Lower bound is due to the presence of a domain shift when the model is trained and tested on different source (K19) and target domains (K16). We observe that the proposed method which generated 6 augmented images for each input image, outperforms the ImageNet baseline by 22% in terms of accuracy. Whereas it performed 20% and 12% better as compared to MocoV2 [22] and InfoMin [23]. IMPaSh [12] is a combination of InfoMin [23] and PatchShuffling augmentations, and our methods still outperform it by 1% while being less computationally expensive. To summarise, the proposed workflow leveraged stain augmentations to encourage the DL model to learn stain and domain-invariant feature representations and outperformed other state-of-the-art methods which begs the question "Is Stain Augmentation really all you need for Domain Generalisation?" Our future work will include automating the selection of the optimal number of augmentations.

TABLE 1: Experimental Results between source (K19) and target (K16) domains

| Method | Training Dataset | Accuracy | Recall | Precision | F1-Score |
|---|---|---|---|---|---|
| ImageNet - Upper Bound | K16 (Target) | 0.942 | 0.942 | 0.941 | 0.941 |
| ImageNet - Lower Bound | K19 | 0.654 | 0.654 | 0.741 | 0.626 |
| SN Macenko [19] | K19 | 0.660 | 0.660 | 0.683 | 0.645 |
| SN Vahadane [18] | K19 | 0.683 | 0.683 | 0.696 | 0.656 |
| InsDis [20] | K19 | 0.694 | 0.694 | 0.766 | 0.659 |
| PIRL [21] | K19 | 0.818 | 0.818 | 0.853 | 0.812 |
| MocoV2 [22] | K19 | 0.675 | 0.675 | 0.816 | 0.642 |
| InfoMin [23] | K19 | 0.750 | 0.750 | 0.824 | 0.752 |
| IMPaSh [12] | K19 | 0.868 | 0.868 | **0.887** | 0.865 |
| Proposed Method | K19 | **0.878** | **0.878** | **0.887** | **0.877** |